\documentclass[10pt,twocolumn,letterpaper]{article}

\usepackage{iccv}
\usepackage{times}
\usepackage{epsfig}
\usepackage{graphicx}
\usepackage{amsmath}
\usepackage{amssymb}

\usepackage[pagebackref=true,breaklinks=true,letterpaper=true,colorlinks,bookmarks=false]{hyperref}

\newcommand{\ADD}[1]{#1}
\newcommand{\RM}[1]{}

\iccvfinalcopy % *** Uncomment this line for the final submission

\def\iccvPaperID{3059}  % *** Enter the ICCV Paper ID here

\ificcvfinal\fi

\usepackage{xspace}

\newcommand{\etals}{\mbox{\emph{et al.\ }'s }}

\newcommand{\LM}[1]{}
\newcommand{\MG}[1]{}
\newcommand{\MA}[1]{}

\begin{document}

%%%%%%%%% TITLE
\title{A Dataset of Multi-Illumination Images in the Wild}

% \author{Lukas Murmann\textsuperscript{1}
%    %
%    \and 
%    Micha\"el Gharbi\textsuperscript{1, 2}
%    \and
%    Miika Aittala\textsuperscript{1}
%    \and
%    Fr\'edo Durand\textsuperscript{1}
%    \\\and
%    MIT CSAIL\textsuperscript{1} \and Adobe\textsuperscript{2} 
% }

\author{
Lukas Murmann\textsuperscript{1}
\and
Micha\"el Gharbi\textsuperscript{1, 2}
\and
Miika Aittala\textsuperscript{1}
\and
Fr\'edo Durand\textsuperscript{1}
\and
\and
\and
MIT CSAIL\textsuperscript{1}
\and
Adobe Research\textsuperscript{2}
}
\maketitle
%\thispagestyle{empty}

%%%%%%%%% ABSTRACT
% 
\begin{abstract}
Collections of images under a single, uncontrolled
illumination~\cite{russakovsky2015} have enabled the rapid advancement of core
computer vision tasks like classification, detection, and
segmentation~\cite{krizhevsky2012,sermanet2013,he2017}. 
But even with modern learning techniques, many inverse problems involving
lighting and material understanding remain too severely ill-posed to be solved
with single-illumination datasets. The data simply does not contain the
necessary supervisory signals.
Multi-illumination datasets are notoriously hard to capture, so the data is
typically collected at small scale, in controlled environments, either using
multiple light sources~\cite{debevec2000, xu2018}, or robotic
gantries~\cite{dana1999,holroyd2010}.
This leads to image collections that are not representative of the variety and
complexity of real-world scenes. 
We introduce a new multi-illumination dataset of more than 1000 real scenes, each
captured in high dynamic range and high resolution, under 25 lighting conditions. 
We demonstrate the richness of this dataset by training state-of-the-art models
for three challenging applications: single-image illumination estimation, image
relighting, and mixed-illuminant white balance.

\end{abstract}

%%%%%%%%% BODY TEXT
\begin{figure}
   \centering
   \includegraphics[width=243pt]{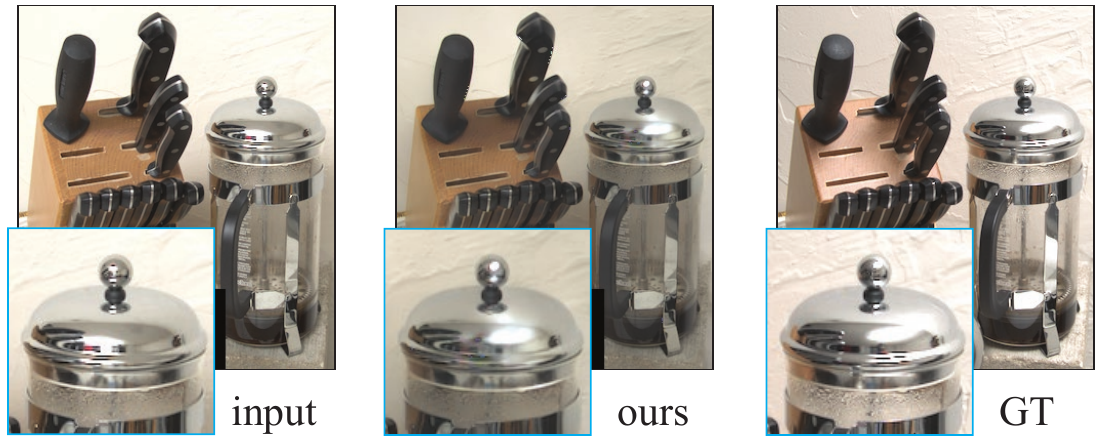}
   \caption{Using our multi-illumination image dataset of over 1000 scenes, we 
      \ADD{can train neural networks to solve challenging vision tasks}. 
      \ADD{For instance, one of our models can} relight an input image to \ADD{a
         novel light direction}. Specular highlights pose a
      significant challenge for many relighting algorithms, but are handled
      gracefully by our network. Further analysis is presented in Section~\ref{sec:relighting}.}\label{fig:teaser}
\end{figure}

\section{Introduction}\label{sec:introduction}

The complex interplay of materials and light is central to the appearance of
objects and to many areas of computer vision, such as inverse problems and
relighting. We argue that research in this area is limited by the scarcity of
datasets --- the current data is often limited to individual samples captured in a
lab setting, e.g.~\cite{dana1999,holroyd2010}, or to 2D photographs that do not
encode the variation of appearance with respect to light~\cite{russakovsky2015}.
While setups such as light stages, e.g.~\cite{debevec2000}, can capture objects
under varying illumination, they are hard to move and require the acquired object
to be fully enclosed within the stage. This makes it difficult to
capture everyday objects in their real environment. 

In this paper, we introduce a new dataset of photographs of indoor surfaces
under varying illumination. Our goal was to capture small scenes at
scale (at least 1,000 scenes). We wanted to be able to bring the capture
equipment to any house, apartment or office and record a scene in minutes. For
this, we needed a compact setup. This appears to be at odds with the requirement
that scenes be illuminated from different directions, since designs such as the
light stage~\cite{debevec2000} have required a large number of individual lights
placed around the scene. We resolved this dilemma by using indirect illumination
and an electronic flash mounted on servos so that we can control its direction.
As the flash gets rotated, it points to a wall or ceiling near the scene, which
forms an indirect ``bounce'' light source. The reflected light becomes the
primary illumination for the scene. We also place a chrome and a gray sphere in
the scene as ground truth measurements of the incoming illumination. 

Our capture process takes about five minutes per scene and is fully automatic.
We have captured over a thousand scenes, each under 25 different illuminations
for a total of 25,000 HDR images. Each picture comes segmented and labeled
according to material. To the best of our knowledge, this is the first dataset
of its kind: offering both everyday objects in context and lighting variations. 

In Section~\ref{sec:application}, we demonstrate the generality and usefulness
of our dataset with three learning-based applications: 
predicting the environment illumination, relighting single images, and correcting inconsistent white
balance in photographs lit by multiple colored light sources.

We release the full dataset, along with our set of tools for processing and
browsing the data, as well as training code and models.

\section{Related Work}\label{sec:related}

\subsection{Multi-Illumination Image Sets}\label{ssec:related_multilum}

Outdoors, the sky and sun are natural sources of illumination varying over time.
Timelapse datasets have been harvested both ``in the wild'' from web
cameras~\cite{weiss2001,sunkavalli2008} or video
collections~\cite{shih2013_timelapse}, or using controlled camera
setups~\cite{stumpfel2004,lalonde2014,laffont2014_outdoor}. 

Indoor scenes generally lack readily-available sources of illumination that
exhibit significant variations. Some of the most common multi-illumination image
sets are collections of flash/no-flash pairs~\cite{petschnigg2004,eisemann2004,aksoy2018}. These image pairs can be
captured relatively easily in a brief two-image burst and enable useful
applications like denoising, mixed-lighting white balance~\cite{hui2017}, or
even BRDF capture~\cite{aittala2015}. Other applications, such as photometric
stereo~\cite{woodham1980} or image relighting~\cite{debevec2000,xu2018}, require more
than two images for reliable results.

Datasets with more than two light directions are often acquired using complex
hardware setups and multiple light sources~\cite{debevec2000,holroyd2010}. A
notable exception, Mohan~\etal~\cite{mohan2007} proposed a user-guided lighting
design system that combines several illuminations of a single object. Like us,
they acquire their images using a stationary motor-controlled light source and
indirect bounce illumination, although within a more restrictive setup, and at a
much smaller scale. For their work on user-assisted image compositing
Boyadzhiev~\etal~\cite{boyadzhiev2013} use a remote-controlled camera and
manually shine a hand-held flash at the scene. This approach ties down the
operator and makes acquisition times prohibitive (they report 20 minutes per
scene). Further, hand-holding the light source makes multi-exposure HDR capture
difficult. In contrast, our system, inspired by work of
Murmann~\etal~\cite{murmann2016}, uses a motor-controlled bounce flash, which
automates the sampling of lighting directions and makes multi-exposure HDR
capture straightforward.

\subsection{Material Databases}\label{ssec:materials}

To faithfully acquire the reflectance of a real-world surface, one typically
needs to observe the surface under multiple lighting conditions. The gold
standard in appearance capture for materials is to exhaustively illuminate the
material sample and photograph it under every pair of viewpoint and light direction,
tabulating the result in a Bidirectional Texture Function (BTF). 
The reflectance values can then be read off this large
table at render-time~\cite{dana1999}. 

A variety of BTF datasets have been
published~\cite{dana1999,lawrence2006,weinmann2014}, but the total number of
samples falls far short of what is typically required by contemporary
learning-based algorithms.
A rich literature exists on simple, light-weight hardware capture
systems~\cite{guarnera2017}, but the corresponding public datasets also 
typically contain less than a few dozen examples.
Additionally, the scope, quality and format of these scattered and small
datasets varies wildly, making it difficult to use them in a unified manner.
Our portable capture device enables us to capture orders of magnitude
more surfaces than existing databases and we record entire scenes  at once
---rather than single objects--- ``in the wild'', outside the laboratory.

Bell~\etal~\cite{bell2013,bell2015material} collected a large dataset of very loosely controlled
photographs of materials from the Internet, enriched with crowd-sourced
annotations on material class, estimated reflectance, planarity and other
properties. 
Inspired by their approach, we collect semantic material class segmentations for
our data, which we detail in section~\ref{ssec:annotation}.
Unlike ours, their dataset does not contain lighting variations.

\ADD{Previous works have investigated the use of synthetic image datasets for
material estimation~\cite{park2018photoshape,weinmann2014material}. But even
carefully crafted synthetic datasets typically do not transfer well to real
scenes due remaining differences in scene complexity, object appearance, and
image formation~\cite{richter2016playing}.}

% I've toned down what kind of contribution we claim.}

%%%%%%%%%% CANDIDATE FOR DELETION - maybe too materials focused for ICCV %%%%%%%%%%
% Several recent learning-based works tackle the difficulty of obtaining
% real-world training data by instead using databases of artist-generated
% synthetic materials \cite{li2017,deschaintre2018}, which are easier to author
% and edit, but on the other hand inherit any shortcomings and biases of the
% artistic content creation process. Li et al.~\cite{li2017} augment their small
% synthetic dataset with loosely labeled real-world data from OpenSurfaces
% \cite{bell2013} using semi-supervised learning techniques. 
%%%%%%%%%% ENDCANDIDATE FOR DELETION %%%%%%%%%%

\section{Dataset}\label{sec:dataset}

\begin{figure*}
  \centering
  \includegraphics[width=0.99\linewidth]{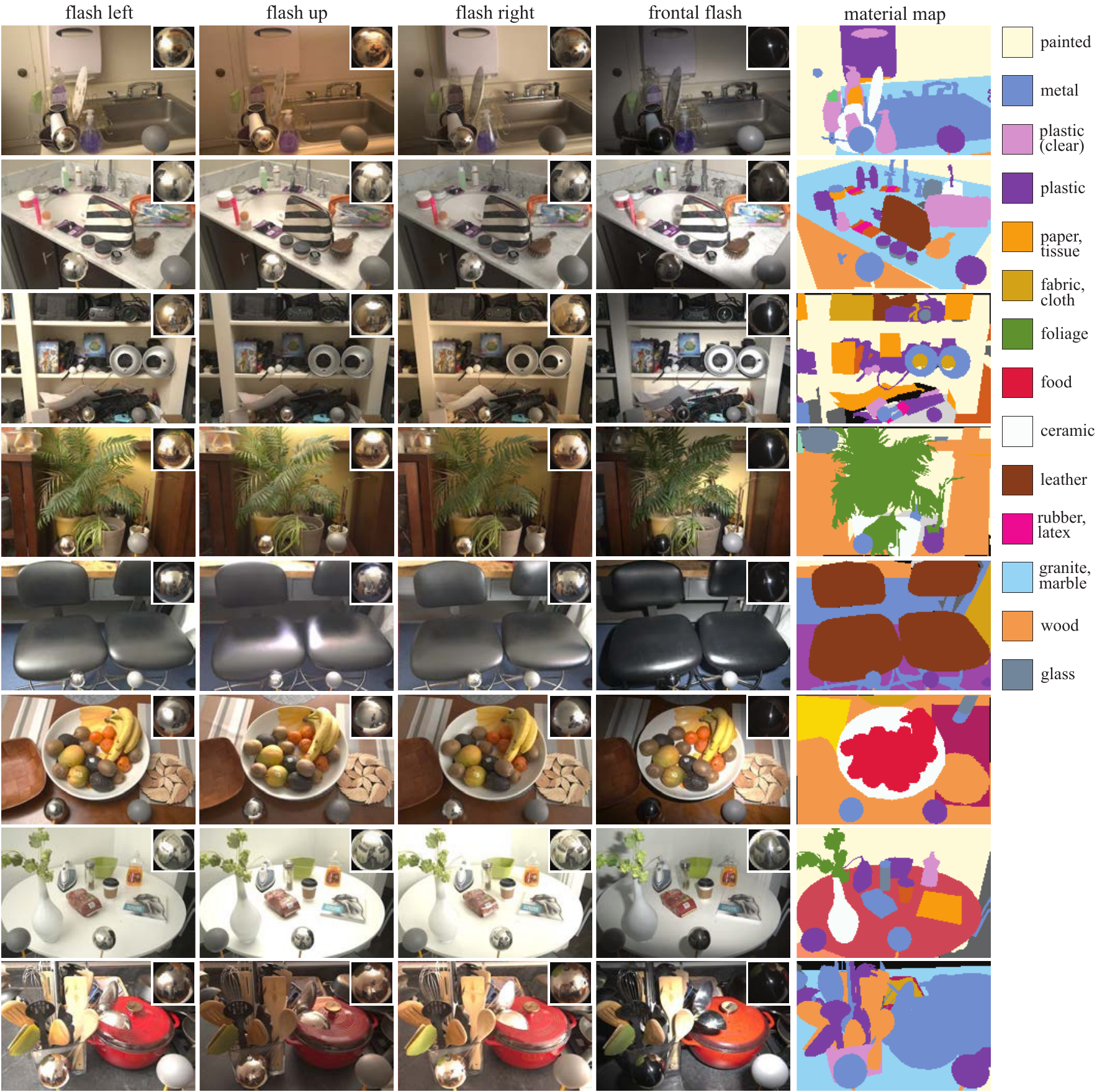} 
  \caption{Eight representative scenes from our dataset. 
    Each scene is captured under 25 unique light directions, 4 of which are 
    shown in the figure. We strived to include a variety of room and material
    types in the dataset. Material types are annotated using dense segmentation
    masks which we show on the right.}\label{fig:tableau}
\end{figure*}

Our dataset consists of 1016 interior scenes, each photographed under 25
predetermined lighting directions, sampled over the upper hemisphere relative to
the camera. The scenes depict typical domestic and office environments. To
maximize surface and material diversity, we fill the scenes with miscellaneous
objects and clutter found in our capture locations. A selection of scenes
is presented in Figure~\ref{fig:tableau}.

In the spirit of previous works~\cite{mohan2007,murmann2016}, our lighting
variations are achieved by directing a concentrated flash beam towards the walls
and ceiling of the room. The bright spot of light that bounces off the wall
becomes a virtual light source that is the dominant source of
illumination for the scene in front of the camera. 

\begin{figure}
  \centering
  \includegraphics[width=243pt]{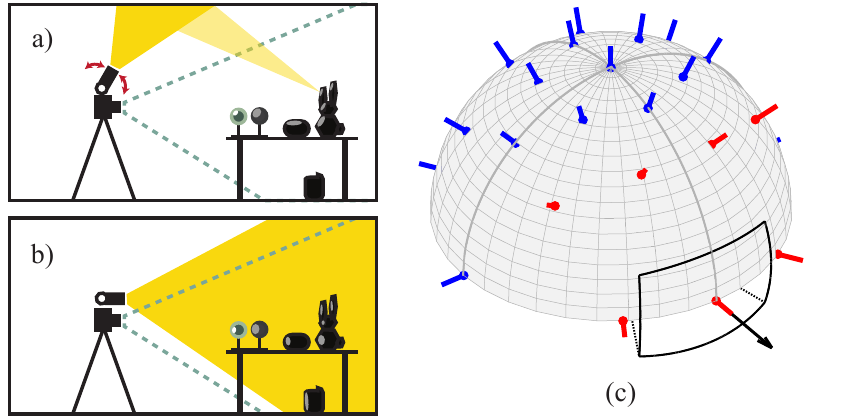} 

  \caption{a) Most of our photographs are lit by pointing a flash unit towards
  the walls and the ceiling, creating a virtual bounce light source that
  illuminates the scene directionally. b) Some of the photographs are captured
  under direct flash illumination, where the beam of light intersects the field
  of view of the camera. c) The flash directions used in our dataset, relative
  to the camera viewing direction and frustum (black). The directions where
  direct flash illumination is seen in the view are shown in red, and the fully
  indirect ones in blue.}
  \label{fig:bounce_flash}
\end{figure}

We can rapidly and automatically control the approximate position of the  bounce
light simply by rotating the flash head over a standardized set of directions
(Figure~\ref{fig:bounce_flash}). This alleviates the need to re-position a
physical light source manually between each exposure~\cite{boyadzhiev2013,masselus2002}.
Our camera and flash system is more portable than dedicated light sources, which
simplifies its deployment ``in the wild''. 

The precise intensity, sharpness and direction of the illumination resulting
from the bounced flash depends on the room geometry and its materials. We record
these lighting conditions by inserting a pair of light probes, a reflective
chrome sphere and a plastic gray sphere, at the bottom edge of every image
\cite{debevec1998}. In order to preserve the full dynamic range of the light
probes and the viewed scene, all our photographs are taken with bracketed
exposures.

As a post-process, we annotate the light probes, and collect dense
material labels for every scene using crowd-sourcing, as described in Section
\ref{ssec:annotation}.

\subsection{Image Capture}\label{ssec:hardware}

Our capture device consists of a mirrorless camera (\emph{Sony $\alpha$6500}),
and an external flash unit (\emph{Sony HVL-F60M}) which we equipped with two
servo motors. \ADD{The servos and camera} are connected to a laptop,
which automatically aims the flash and fires the exposures in a pre-programmed
sequence. The 24mm lens provides a $52^{\circ}$ horizontal and $36^{\circ}$
vertical field of view.

At capture time, we rotate the flash in the 25 directions depicted in
Figure~\ref{fig:bounce_flash}, and capture a 3-image exposure stack for each
flash direction.
We switch off any room lights and shut window blinds, which brings the average
intensity of the ambient \ADD{light} to less than 1\% of the intensity of the
flash \ADD{illumination}.
For completeness, we capture an extra, ambient-only, exposure stack with
the flash turned off. 

The 25 flash directions are evenly spaced over the upper hemisphere. In 18 of
these directions, the cone of the flash beam falls outside the view of the camera,
and consequently, the image is only lit by the secondary illumination from the
bounce. In the remaining 7 directions, part or all of the image is lit by the
flash directly. In particular, one of the directions corresponds to a typical
frontal flash illumination condition.

Capturing a single set (78 exposures) takes about five minutes with our current
setup. The capture speed is mostly constrained by the flash's recycling time
(around 3.5 seconds at full power). Additional battery extender packs or
high-voltage batteries can reduce this delay for short bursts. We found them
less useful when capturing many image sets in a single session, where heat
dissipation becomes the limiting factor.

\subsection{HDR processing}
The three exposures for each light direction are 
\ADD{bracketed in 5-stops increments} to avoid clipped highlights and excessive noise in
the shadows. The darkest frame is exposed at $f/22$~ISO100, the middle exposure is
$f/5.6$~ISO200, and the brightest image is recorded at $f/5.6$~ISO6400. 
The shutter speed is kept at the camera's fastest flash sync time, 
$1/160^\text{th}$ second to minimize ambient light.
The camera sensor has 13 bits of useful dynamic range at ISO100 (9 bits at
ISO6400).
Overall, our capture strategy allows us to reconstruct HDR images with at least
20 bits of dynamic range. 

Using the aperture \ADD{setting to control} exposure bracketing could lead
to artifacts from varying defocus blur. We limit this effect by manually focusing
the camera to the optimal depth, and by avoiding viewpoints with depth complexity
beyond the depth-of-field that is achieved at $f/5.6$. 

After merging exposures, we normalize the brightness of the HDR image by
matching the intensity of the diffuse gray sphere. The gray sphere also serves
as a reference point for white balance. This is especially useful in
brightly-colored rooms that could otherwise cause color shifts. 

\subsection{Dataset Statistics}\label{ssec:statistics}\label{ssec:annotation}
To ensure our data is representative of many real-world scenes, we collected images in 95
different rooms throughout 12 residential and office buildings, which allowed
us to capture a variety of materials and room shapes. 

\begin{figure}
  \centering
  \includegraphics[width=243pt]{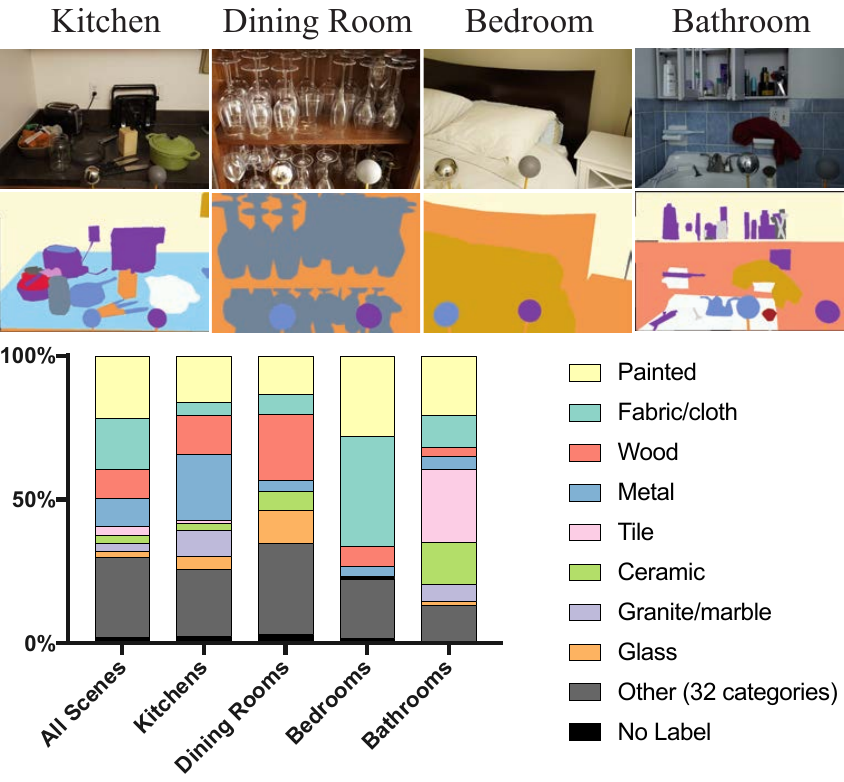} 
  \caption{Crowd-sourced material annotations show that painted surfaces, fabrics,
  wood, and metal are the most frequently occurring materials in our dataset,
  covering more than 60\% of all pixels. For some room types, the material
  distribution is markedly different from the average. For example, in kitchens
  we frequently encounter wood (shelves) and metal (appliances), bedroom scenes
  show a high frequency of fabrics, and the material distribution of bathrooms
  is skewed towards tiles and ceramics.}\label{fig:room_type_barchart}
\end{figure}

In order to analyze the materials found throughout our dataset, we obtain dense
material labels segmented by crowd workers, as shown in
Figure~\ref{fig:tableau} and~\ref{fig:room_type_barchart}. These annotations are
inspired by the material annotations collected by Bell~\etal~\cite{bell2013},
whose publicly available source code forms the basis of our annotation pipeline.

Figure \ref{fig:room_type_barchart} shows the distribution of materials in our
data set. Specific room types have material distributions that differ markedly
from the unconditioned distribution. For example, in kitchens we frequently find
metal and wooden surfaces, but few fabrics (less than 5\% of pixels). Bedrooms
scenes on the other hand show fabrics in 38\% of the pixels, but contain almost
no metal surfaces (less than 4\%).

\section{Applications}\label{sec:application}
In this section we present learning-based solutions to three long-standing
vision problems: single-image lighting estimation, single-image relighting and
mixed-illuminant white-balance.
Our models are based on standard convolutional architectures, such as the
U-net~\cite{ronneberger2015_unet}. For all experiments, we normalize the
exposure and white balance of our input images with respect to the gray sphere.
We also mask out the chrome and gray spheres with black squares both at training
and test time to prevent the networks from using this information directly.

   \subsection{Predicting Illumination from a Single Image}\label{sec:probe_prediction}
   Single-frame illumination estimation is a challenging problem that arises e.g.\
when one wishes to composite a computer-generated object into a real-world
image~\cite{debevec1998}. Given sufficient planning (as is common for visual
effects in movies), illumination can be recorded at the same time the backdrop
is photographed (e.g.\ by using a light probe). This is rarely the case for a
posteriori applications. In particular, with the growing interest in augmented
reality and mixed reality, the problem of estimating the illumination in
uncontrolled scenes has received increased attention.

Several methods have explored this problem for outdoor
images~\cite{lalonde2012,georgoulis2016,rematas2017,hold-geoffroy2018,lombardi2016} as well
as indoor environments~\cite{gardner2017}.
Noting the lack of viable training data for indoor scenes, Gardner~\etal
explicitly detect light sources in LDR panoramas~\cite{xiao2012}. 
Our proposed dataset includes HDR light probes in every scene which makes it
uniquely suitable for illumination prediction and other inverse rendering
tasks~\cite{barron2015} in indoor environments.

\subsubsection{Model}
We approach the single image illumination prediction problem by training a
convolutional network on $256\times 256$ image crops from our dataset. 
We ask the network to predict a $16\times 16$ RGB chrome sphere, that we compare
to our ground truth probe using an $L_2$ loss. \RM{Both the input and the output are
in the log-domain to limit the dynamic range of the data passing through
the network.}
The $256\times 256$ input patch is processed by a sequence of convolution, ReLU,
and Max-pooling layers, where we halve the spatial resolution and double the
number of feature maps after each convolution. When the spatial resolution
reaches $1\times1$ pixel, we apply a final, fully-connected layer to predict 768
numbers: these are reshaped into a $16\times 16$ RGB light probe image.
Exponentiating this images yields the final, predicted environment map.
We provide the network details in supplemental material.

\subsubsection{Compositing synthetic objects}

\begin{figure}
  \centering
  \includegraphics[width=248pt]{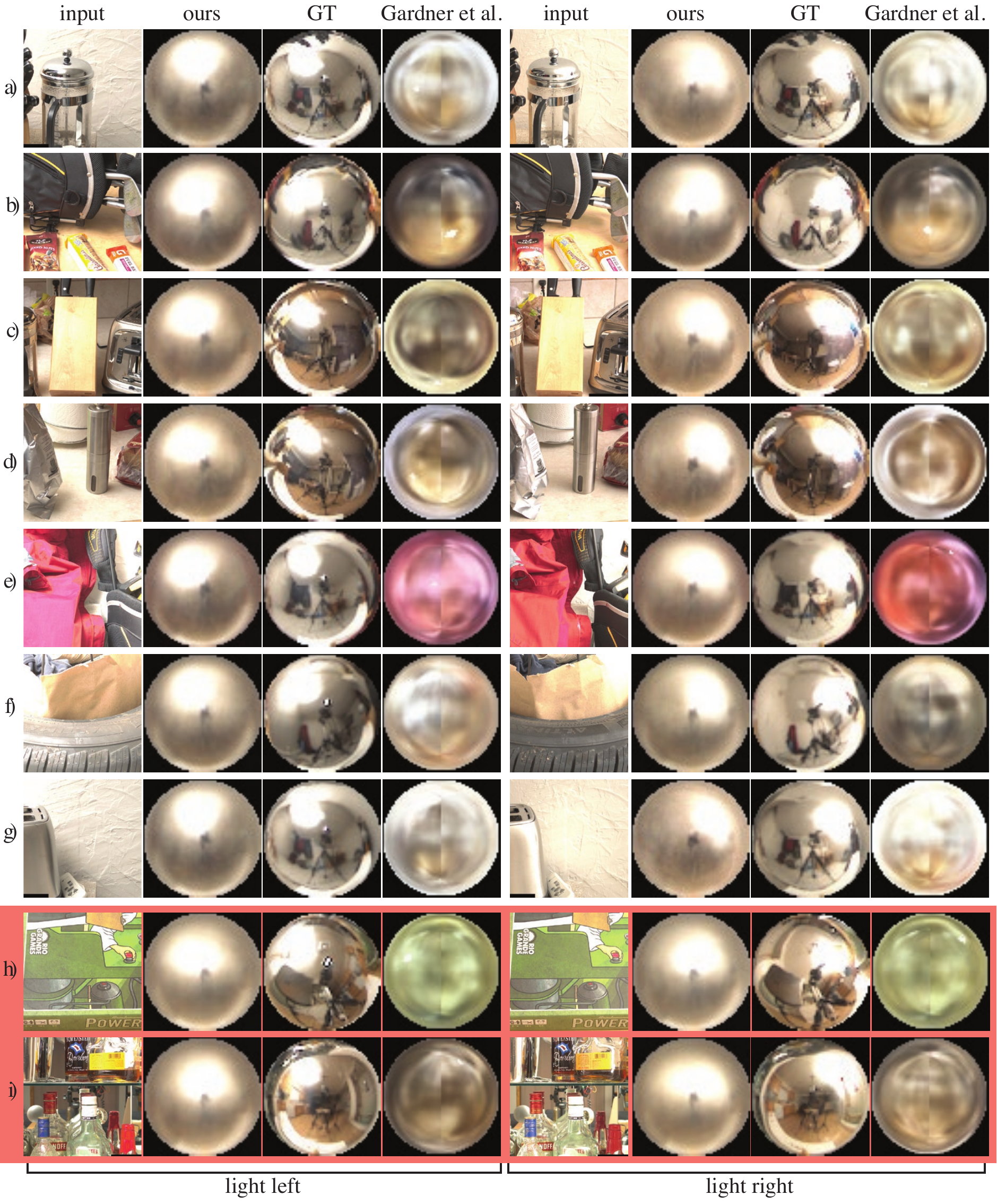}
  \caption{As the first application of our \ADD{dataset}, we train a deep network to
  predict environment maps from single input images. Our model consistently predicts
  the dominant light direction of the ground truth environment map. The model
  successfully estimates illumination based on shiny objects (a and g) 
  and diffuse reflectors (e.g. row f). Rows h) and i) show failure cases where the
  network predicts low-confidence outputs close to the mean direction.
  We compare to
  Gardner~\etals algorithm~\cite{gardner2017} which, while predicting visually
  plausible environment maps, lacks the precise localization of highlights shown
  by our technique.
  (Please ignore the vertical seam in Gardner~\etal's result. Their model uses
  a differing spherical parametrization, which we remap to our coordinate system
  for display.)}\label{fig:chromeball_predict_a}
\end{figure}

\begin{figure}[!bt]
  \centering
  \includegraphics[width=\linewidth]{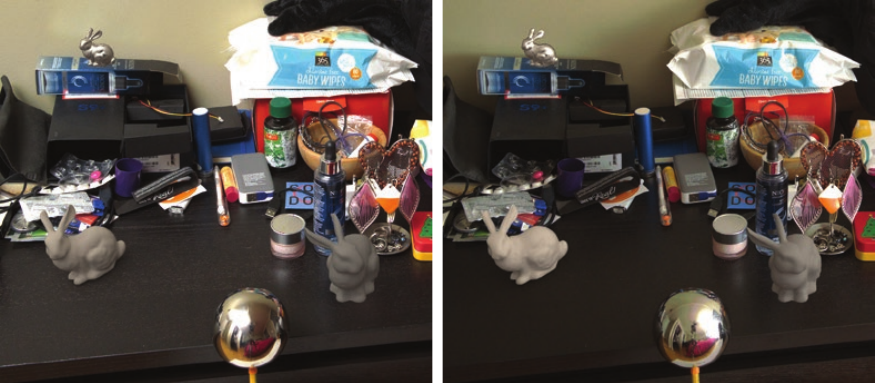} 
  \caption{We use the environment maps predicted by our model to illuminate
    virtual objects and composite them onto one of our test scenes. The light probe in the bottom of the frame shows the ground truth lighting. (Note that these probes are masked out before feeding the image to the network).}
  \label{fig:bunnies}
\end{figure}

Figure~\ref{fig:chromeball_predict_a} shows some compositing results on a held-out
test set. While our model does not always capture the finer color
variations of the diffuse global illumination, its prediction of the dominant
light source is accurate. Figure~\ref{fig:bunnies} shows one of the
test scenes, with synthetic objects composited. The synthetic geometry is
illuminated by our predicted environment maps and rendered with a path-tracer.
Note that the ground truth light probes visible in the figure were masked during
inference, and therefore \emph{not} seen by our network.

\subsubsection{Evaluation}

We evaluated our model on a held-out test subset of our data and compared it to
a state-of-the-art illumination prediction algorithm
by Gardner~\etal~\cite{gardner2017}.
Compared to their technique, our model more accurately predicts 
the direction of the bounce light source (see Figure~\ref{fig:chromeball_predict_a}).
In comparison, Gardner~\etal's model favors smoother environment maps and is
less likely to predict the directional component of the illumination.
For visual comparison, we warp the $360^\circ$ panoramas produced by
their technique to the chrome \ADD{sphere} parameterization that is used throughout
our paper.

In order to quantify the performance of the chrome \ADD{sphere} predictor,
\ADD{we  analyzed the angular distance between the predicted and true center of
the light source for 30 test scenes. Our technique achieves a mean angular
error of $26.6^\circ$ (std.\ dev.\ $10.8^\circ$), significantly outperforming
Gardner~\etal's method, which achieves a mean error of $68.5^\circ$ (std.\
dev.\ $38.4^\circ$).}
Visual inspection suggests 
that the remaining failure cases of our technique are due to left/right symmetry
of the scene geometry, mirrors, or simply lack of context in the randomly chosen
input crops (see Figure~\ref{fig:chromeball_predict_a} bottom).

\begin{figure}
  \centering
  \includegraphics[width=223pt]{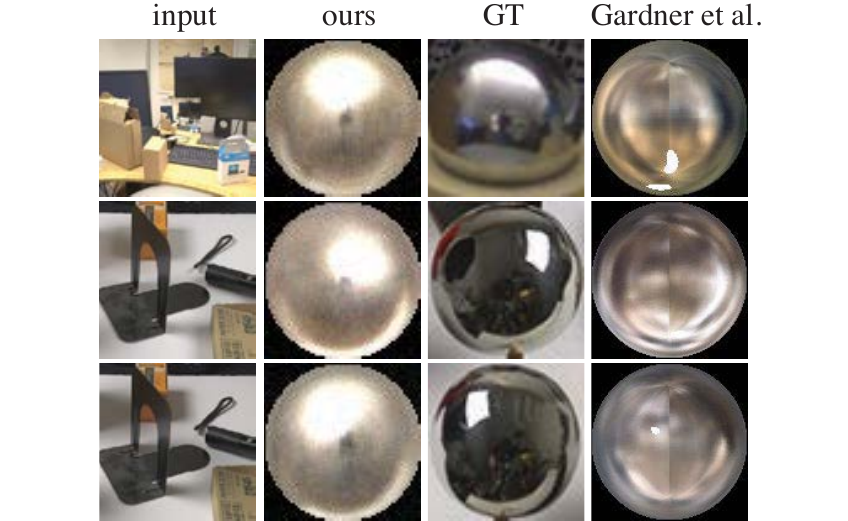}

  \caption{We validate our model's ability to generalize beyond
  bounce flash illumination. The top row show an office
  scene with regular room lights. The bottom two rows show a scene
  illuminated by softboxes, first lit from the right and then 
  from the left. The second set of results suggests that our model
  can aggregate information from shadows to infer the light source position.}
  \label{fig:predict_real_lighting}
\end{figure}

We verified \ADD{that} our model \ADD{generalizes} beyond bounce flash light
sources using a small test set of pictures taken under general non-bounce flash
illumination. The results of this experiment are presented in
Figure~\ref{fig:predict_real_lighting}.

   \subsection{Relighting}\label{sec:relighting}
   A robust and straightforward method to obtain a relightable model of a scene
is to capture a large number of basis images under varying illumination, and
render new images as linear combinations of the basis elements. Light
stages~\cite{debevec2000} work according to this principle. 
With finitely many basis images, representing the high frequency content of a
scene's light transport operator (specular highlights, sharp shadow
boundaries, etc.) is difficult.
Despite this fundamental challenge, prior work has successfully exploited the
regularity of light transport in natural scenes to estimate the transport
operator from sparse samples~\cite{nayar2004,peers2009,wang2009}.
Recent approaches have employed convolutional neural networks for the task,
effectively learning the regularities of light transport from synthetic training
data and reducing the number of images required for relighting to just a handful
\cite{xu2018}. 

In our work, we demonstrate relighting results from a \emph{single
input image} on real-world scenes that exhibit challenging phenomena, such as
specular highlights, self-shadowing, and interreflections.

\subsubsection{Model}

We cast single-image relighting as an image-to-image translation problem.
We use a convolutional neural network based on the
U-net~\cite{ronneberger2015_unet} architecture to map from images illuminated
from the left side of the camera, to images lit from the right (see
supplemental material for details).
Like in Section~\ref{sec:probe_prediction}, we work in the log-domain to limit
the dynamic range of the network's internal activations.
We use an $L_1$ loss to compare the \emph{spatial gradients} of our relit output
to those of the reference image, lit from the right. 
We found this gradient-domain loss to yield sharper results. It also allows the
network to focus more on fine details without being overly penalized for
low-frequency shifts due to the global intensity scaling ambiguity (the left-
and right-lit images might not have the same average brightness, depending on
room geometry). 

\begin{figure}[!bth]
  \centering
  \includegraphics[width=\linewidth]{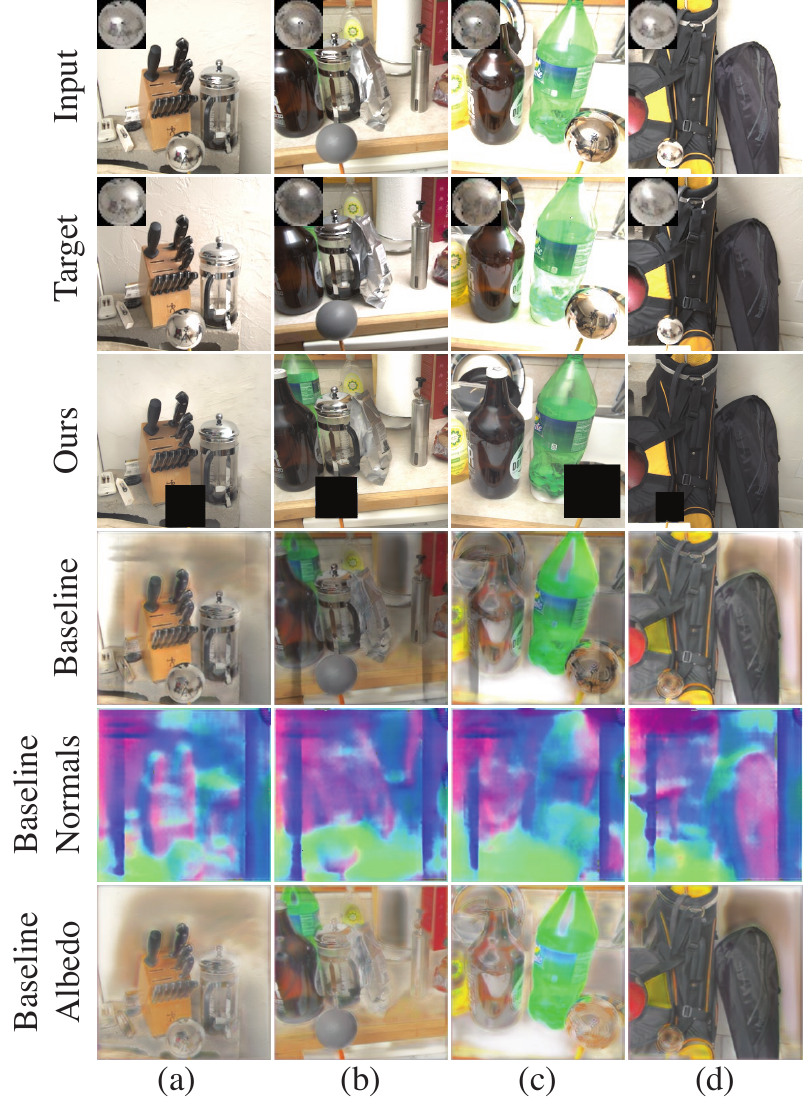} 
  \caption{
  The second application of our data is learning image relighting using a single
  input image. The trained model synthesizes moving specular highlights (a, b,
  c) and diffuse shading (b, d), and correctly renders shadows behind occluders
  (d). For the  baseline result, we first estimate normals and diffuse albedo
  using published models, and then re-render the image as lit by the target
  environment map.}
  \label{fig:relightingcomp}
\end{figure}

\subsubsection{Evaluation}

Our model faithfully synthesizes specular highlights, self- and cast-shadows, as
well as plausible shading variations.
Without a large-scale training dataset for an end-to-end solution, a valid
straw man approach would be to use existing techniques and decompose the input
into components that can be manipulated for relighting (e.g.\ normals and
albedo).
We provide such a baseline for comparison. It uses a combination of deep
single-image normal estimation~\cite{zhang2016physically} and learned intrinsic
image decomposition~\cite{innamorati17decomposing} (see
Figure~\ref{fig:relightingcomp}). 
Both components are state-of-the-art in their respective fields and have source
code and trained models publicly available. 

This baseline illustrates the challenges in decomposing the single-image
relighting problem into separate inverse rendering sub-problems.
\ADD{Specifically, major sources of artifacts include: incorrect or
blurry normals, and incorrect surface albedo due to the overly simplistic
Lambertian shading model.
The normals and reflectance estimation networks were independently trained to
solve two very difficult problems. This is a arguably more challenging than our
end-to-end relighting application and, also unnecessary for plausible
relighting.}

Our end-to-end solution does not enforce this explicit geometry/material
decomposition and yields far superior results. 
More relighting outputs produced by our model are shown in
Figure~\ref{fig:teaser} and in the supplemental material.

   \subsection{Mixed-Illumination White-Balance}
   White-balancing an image consists in neutralizing the color cast caused by
non-standard illuminants, so that the photograph appears lit by a standardized
(typically white) light source. White-balance is under-constrained, and is often
solved by modeling and exploiting the statistical regularities in the colors of
lights and objects. The most common automatic white balance algorithms make the
simplifying assumption that the entire scene is lit by a single illuminant.
See \cite{gijsenij2011computational} for a survey. 
This assumption rarely holds in practice. For instance, an interior
scene might exhibit a mix of bluish light (e.g.\ from sky illuminating the
scene through a window) and warmer tones (e.g.\ from the room's artificial
tungsten light bulbs). 
Prior work has formulated a local gray-world assumption to generalize white
balance to the mixed-lighting case~\cite{ebner2003_wb_local_retinex}, exploiting
the difference in light colors in shadowed vs.\ sunlit areas for outdoor scenes
\cite{kawakami2005_wb}, or flash/no-flash image
pairs~\cite{matsuoka2015_wb,hui2016_wb,hui2017_illuminant}.
%
% \ADD{Hsu~\etal~\cite{hsu2008_wb} assume the scene is lit by two illuminants whose
% colors are provided by the user. Boyadzhiev~\etal~\cite{boyadzhiev2012_wb}
% require the user to annotate image areas neutral or uniformly lit objects with
% scribbles.}

Here again, we approach white-balancing as a supervised learning problem.
Because our dataset contains high-dynamic range linear images with multiple
lighting conditions, it can be used to simulate a wide range of new mixed-color
illuminations by linear combinations. 
We exploit this property to generate a training dataset for a neural network
that removes inconsistent color casts from mixed-illumination photographs.

\subsubsection{Mixed-illuminant data generation}

To create a training set of input/output pairs, we extract $256\times256$ patches
from our scenes at multiple scales.
For each patch, we choose a random number of light sources $n\in\{1,\ldots,4\}$.
Each light index corresponds to one of 25 available flash directions, selected
uniformly at random without replacement. We denote by $I_1,\ldots,I_n$ the
corresponding images.

For each light $i$, we sample its color with hue in $[0, 360]$, and saturation
in $[0.5, 1]$, represented as a positive RGB gain vector $\alpha_i$, normalized
such that $||\alpha_i||_1$ = 1.
We randomize the relative power of the light sources by sampling a scalar
exposure gain $g_i$ uniformly (in the log domain) between $-3$ and $+2$ stops.
We finally assemble our mixed-colored input patch as the linear combination:
$I = \frac{1}{n}\sum_{i=1}^{n} \alpha_i g_i I_i$.
%
% \begin{equation}
%   I = \frac{1}{n}\sum_{i=1}^{n} \alpha_i g_i I_i,
% \label{eq:wb_input}
% \end{equation}
%

We define the color-corrected target similarly, but without the color
gains: $ O = \frac{1}{n}\sum_{i=1}^{n} g_i I_i$.
%
% \begin{equation}
% O = \frac{1}{n}\sum_{i=1}^{n} g_i I_i,
% \label{eq:wb_target}
% \end{equation}

% For data augmentation, we randomly flip and rotate our training patches in $90$ degrees increments.

\begin{figure}[!bt]
  \centering
  \includegraphics[width=\linewidth]{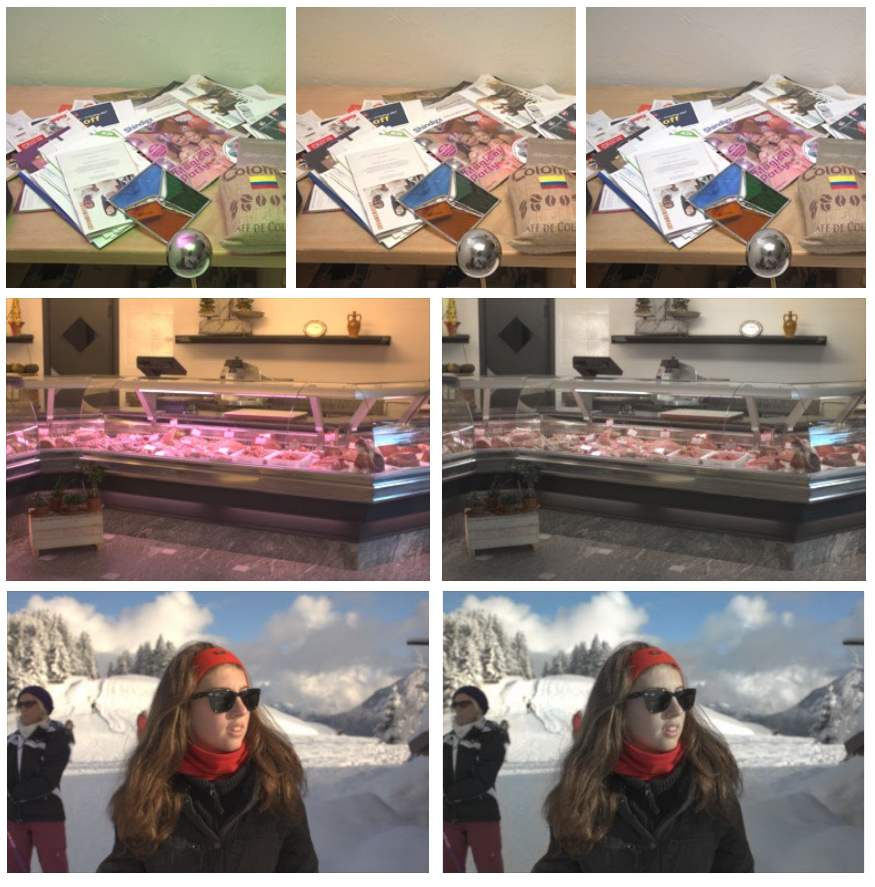} 
  \caption{The first row shows a mixed white-balance result on an image from a
    held-out test set. The input (left) is a linear combination of several
  (two here) illuminations of the same scene under varied color and intensity.
  The reference image (right) has the same energy but no color cast. Our output
  (middle) successfully removes the green and magenta shifts. The simple model
  we trained on our dataset, generalizes well to unseen, real RAW
  images (second row). The most noticeable failure case are skin tones (third
  row), which are entirely absent from our data set of static scenes.}
  \label{fig:wb_result}
\end{figure}

\subsubsection{Model}

Like for the relighting problem, we use a simple convolutional network based on
a U-net~\cite{ronneberger2015_unet} to predict white-balanced images from
mixed-lighting inputs (details in the supplemental).

% Predict log-chroma
To reduce the number of unknowns and alleviate the global scale ambiguity, we
take the log transform of the input and target images, and decompose them in 2
chrominance $u$, $v$, and a luminance component $l$~\cite{barron2015_cc}:
\begin{align}
u &= \log(I^r + \epsilon) -  \log(I^g + \epsilon), \\
v &= \log(I^b + \epsilon) -  \log(I^g + \epsilon), \\
l &= \log(I^g + \epsilon),
\end{align}
where $\epsilon=10^{-4}$, and the superscripts stand for the RGB color channels.

Our network takes as input $u, v, l$ and outputs two correctly white-balanced
chroma components.
We assemble the final RGB output from $l$ and the predicted
chroma, using the reverse transform.
Our model is trained to minimize an $L_2$ loss over the chroma difference.

\subsubsection{Results}

Our model successfully removes the mixed color cast on our test set and
generalizes beyond, to real-world images. 
The main limitation of our technique is its poor generalization to skin tones,
to which the human eye is particularly sensitive, but which are absent from our
dataset of static indoor scenes.
We present qualitative results in Figure~\ref{fig:wb_result} and in the
supplemental video.

% report error in hue degree
% Figure in/ours/groundtruth
% Figure real scenes

\section{Limitations}

A limitation of our capture methodology is that it requires good bounce surfaces
placed not too far from the scene. This precludes most outdoor scenes and large
indoor rooms like auditoriums. Our capture process requires the scene to remain
static for several minutes, which keeps us from capturing human subjects. 
Compared to light stages or robotic gantries, the placement
of our bounce light sources has more variability due to room geometry, and the
bounce light is softer than hard lighting from point light sources.
Finally, we only capture 25 different illuminations, which
is sufficient for \ADD{diffuse} materials but under-samples highly specular ones.

\section{Conclusions}

We have introduced a new dataset of indoor object appearance under 
varying illumination. We have described a novel capture methodology
based on an indirect bounce flash which enables, in a
compact setup, the creation of virtual light sources. 
Our automated capture protocol allowed us to acquire over a 
thousand scenes, each under 25 different illuminations. 
We presented applications in environment map estimation, single-image
relighting, and mixed white balance that can be trained from scratch using our
dataset. We will release the code and data. % under permissive licenses.

\section*{Acknowledgments}
This work was supported in part by
the DARPA REVEAL Program under Contract No. HR0011-
16-C-0030.

%
% \appendix
% \section{Network architectures}
% \input{apdix_networks}\label{sec:net_arch}

{\small
\bibliographystyle{ieee_fullname}
\bibliography{ms}
}

\end{document}

%%%%%%%%% TITLE

      {
   \newpage
   \null
   \begin{center}
      {\Large \bf  {A Dataset of Multi-Illumination Images in the Wild}  \par}
      % additional two empty lines at the end of the title
%       \vspace*{24pt}
%       {
%       \large
%       \begin{tabular}[t]{c}
% Paper ID \cvprPaperID
%       \end{tabular}
%       \par
%       }
%       % additional small space at the end of the author name
%       \vskip .5em
%       % additional empty line at the end of the title block
      \vspace*{6pt}
   \end{center}
   }

\appendix
\section{Crowd-sourced data annotation}

For each scene, we include dense material labels, segmented by crowd workers. 
These annotations are inspired by the material annotations
collected by Bell~\etal~\cite{bell2013}, whose publicly available source code forms the basis
for our data annotation system.
Departing from Bell~\etal, we strive to densely label each scene
with at least 95\% coverage. In comparison, the images published
by Bell~\etal have an average coverage of 20\%.

Annotations with above 95\% coverage
also set our dataset apart from semantic
segmentation datasets such as Coco~\cite{tsungyi2017} 
and Pascal~\cite{everingham2012}, 
which generally exhibit many unlabeled background pixels. 
In particular, the Coco 2014 training set has
an average pixel-level coverage of 29.6\%, while Pascal VOC2012
includes annotations for 25.5\% of the pixels.

%%%% Remove this paragraph so that suppl fits on one page.
% For semantic annotations, the relevant information is usually contained in the
% foreground pixels. For instance, in a picture of a horse (foreground) standing on a meadow (background),
% the horse is arguably more important for semantic learning tasks. For material segmentation however,
% the meadow in the background can contain dirt, grass, rocks, water, a fence made of wood and metal, etc.
% which may give it equal or higher importance when analyzing materials.
%%%%%

In early crowd sourcing experiments, we found it difficult to achieve high
annotation coverage using the 
polygon-based segmentation user interface from Bell~\etal~\cite{bell2013}, which also forms the backbone of the Coco annotations~\cite{tsungyi2017}. %
While segmentation using a single polyline is an effective solution for foreground
objects without holes, encouraging a thorough labeling of the background turned out more difficult. 
We found that background regions must frequently be split into several polygons to avoid
accidentally including foreground objects. Further, we observed that 
labeling both foreground and background would almost double the work required, 
as each occluding contour must be traced twice.

We overcome these limitations by segmenting objects in order, from front to back. 
In a typical segmentation session, a worker starts by labeling
foreground objects that are not occluded (i.e.\ objects closest to the camera).
Once these occluding objects are labeled, the worker can ``extract'' them from the image. 
%\MA{there was talk about adding a figure here but maybe no point. but it's shown in video, do we want to point that out, or is this straightfowrard anyway?} 
As background polygons are automatically masked by extracted foreground shapes,
the worker can then segment the ``second layer'' without worrying that their 
newest polygon might overlap previously segmented areas.
Our front-to-back segmentation interface is efficient since occluding contours
only have to be traced once, background objects are segmented into a single
contiguous polygon, and the front-to-back logic ensures there are neither gaps
nor overlap between foreground and background annotations.

\begin{figure}
  \centering
  \includegraphics[width=243pt]{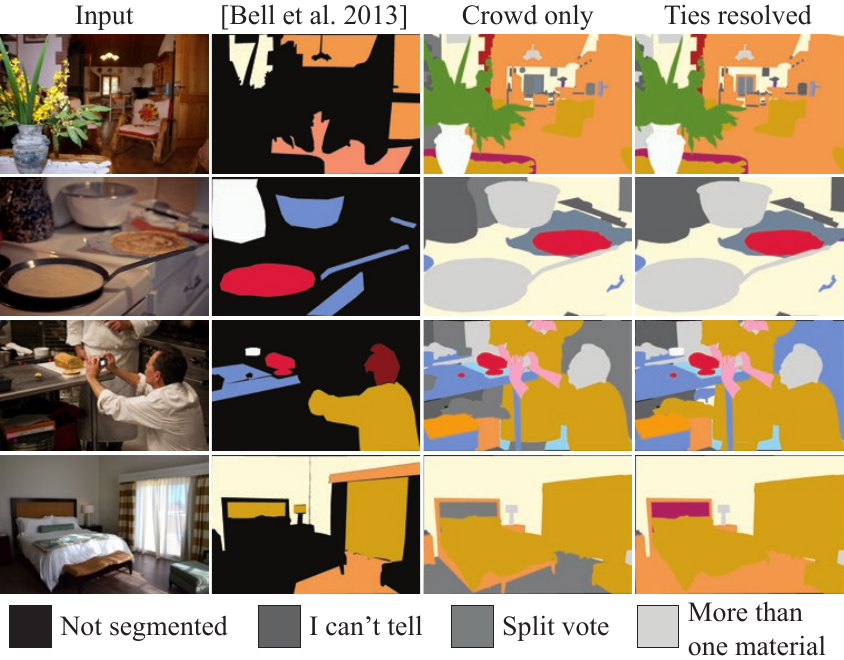} 
  \caption{Obtaining full-image annotations using
  an unmodified version of~\cite{bell2013} is challenging (column 2). Our modified
  version of their UI lets workers segment images front-to-back, which allows background
  objects to be captured in a single contiguous shape (column 3). In some cased (see
  wooden floor in bottom row), crowd-sourced votes are split, which we resolve
  manually for the final result (column 4).
  }
  \label{fig:os_comparison}
\end{figure}

We found that crowd workers on Amazon Mechanical Turk
were able to reliably segment scenes in front-to-back order after viewing a short
tutorial video that introduces our user interface and demonstrates the semantics
of front-to-back ordering. 

After the segmented shapes are submitted to our server, we add them to a work queue where
workers are asked to choose the material for each segment. The choice of material categories
and annotation interface follows Bell~\etal~\cite{bell2013}. Each shape is presented to five
workers and materials
are determined by a simple majority vote. In cases where no material receives a majority,
we resolve ties manually. Figure~\ref{fig:os_comparison} compares the segmentations
obtained using our technique with thouse obtained using the original user interface.
\section{Network architectures}

Let $C_k$ denote a Conv--ReLU layer with $k$ $3\times 3$ kernels.
$P_k$ is a $k\times k$ max-pooling operator,
and $L_k$ is a $1\times 1$ convolution with $k$ linear outputs (no activation).
$U_k$ is a $k\times k$ bilinear upsampling layer.

\paragraph{Illumination prediction}
Our fully convolutional illumination prediction network 
takes $256\times256$ color images as input and produces $16\times 16$ RGB light
probe images (i.e.\ 768 output floats). Its architecture is given by:
$
% 256x256
  C_{32}
  P_{2}
% 128x128
  C_{64}
  P_{2}
% 64x64
  C_{128}
  P_{2}
% 32x32
  C_{256}
  P_{2}
% 16x16
  C_{512}
  P_{4}
% 4x4
  D
  C_{512}
  P_{4}
% 1x1
  C_{512}
  L_{768}.
$

\paragraph{Relighting and white-balance}

Both the relighting and white-balance application use a U-net with 7
downsampling stages and take 3-channel images as input. The relighting output is
a color image (3 channels), but the white-balance model produces only the chroma
components (2 channels), which are then combined with the input luminance.
The U-net encoder can be described as:

$
% 128
  (C_{64})^2
  P_{2}
% 64
  (C_{128})^2
  P_{2}
% 32
  (C_{256})^2
  P_{2}
% 16
  (C_{512})^2
  P_{2}
% 8
  (C_{512})^2
  P_{2}\\
% 4
  (C_{512})^2
  P_{2}
% 2
  (C_{512})^2
  P_{2}
% 1
  (C_{512})^2.
$

And the decoder is given by:

$
  U_{2}
% 2
  (C_{64})^2
  U_{2}
% 4
  (C_{128})^2
  U_{2}
% 8
  (C_{256})^2
  U_{2}
% 16
  (C_{512})^2
  U_{2}\\
% 32
  (C_{512})^2
  U_{2}
% 64
  (C_{512})^2
  U_{2}
% 128
  (C_{512})^2
  L_{2\text{or}3},
$
with additive skip-connections between matching resolutions.

{\small
\bibliographystyle{ieee}
\bibliography{multilum}
}